\documentclass{llncs}
\usepackage[misc,geometry]{ifsym}
\usepackage{graphicx}
\usepackage{aliascnt}
\usepackage{remreset}
\usepackage{hyperref}

\begin{document}
\title{Towards Automated Tuberculosis detection using Deep Learning}

\author{Sonaal Kant\textsuperscript{*}\inst{1}\\ 
\scriptsize \email{sonaal@paralleldots.com}\\
\normalsize Muktabh Mayank Srivastava\thanks{Authors contributed equally.}\inst{1}\\
\scriptsize \email{muktabh@paralleldots.com}\\
}

\institute{Paralleldots, Inc.\inst{1}}
\maketitle


\begin{abstract}
Tuberculosis(TB) in India is the world's largest TB epidemic \cite{world2009global}. TB leads to 480,000 deaths every year \cite{NSP2017}. Between the years 2006 and 2014, Indian economy lost US\$340 Billion due to TB. This combined with the emergence of drug resistant bacteria in India makes the problem worse \cite{TBWSJ}. The government of India has hence come up with a new strategy which requires a high-sensitivity microscopy based TB diagnosis mechanism \cite{NSP2017}. \\
We propose a new Deep Neural Network based drug sensitive TB detection methodology with recall and precision of 83.78\% and 67.55\% respectively for bacillus detection. This method takes a microscopy image with proper zoom level as input and returns location of suspected TB germs as output. The high accuracy of our method gives it the potential to evolve into a high sensitivity system to diagnose TB when trained at scale. \\
\end{abstract}

\section{Introduction}

Tuberculosis(TB) is a major public health problem in the world, claiming a life every 20 seconds, and India has a quarter of the world's cases \cite{world2009global}. TB used to be curable since 1940s, but things started getting bad with the emergence of Drug Resistant TB variants like Multi Drug Resistant, eXtensive Drug Resistant and Total Drug Resistant form of the TB pathogen. As of 2012, India's Drug Resistant TB rate is much higher than the WHO's reported 2\%-3\% \cite{TBWSJ}. A large undernourished population with weak immunity, latent cases due to unrestrained spread, high cost of treatment for drug resistant forms and densely populated urban clusters are some reasons for the current situation in India. India has adopted a new policy \cite{NSP2017} which aims to diagnose TB early and treat it properly, thus, preventing the spread. This new strategy has four strategic pillars of DTPB (Detect-Treat-Prevent-Build). The Detect pillar involves scaling up of free high-sensitivity  diagnostic tests and algorithms. Sputum Smear Microscopy is used for primary diagnosis of drug sensitive TB. However, with 20 million Indian TB symptomatics undergoing the microscopy test every year, the resources are burdened. 

The work at a TB diagnosis center given a sputum slide can be divided into two parts: 1. Focus the microscope on slides to a level on which bacilli become visible 2. Search and locate bacilli in the slide to estimate the extent of infection. In this work we propose an automated deep learning based method and develop a prototype for locating TB bacteria with high precision and recall given a microscope view-field with a proper zoom level. Automation of the task of locating bacilli reduces the work at a diagnosis facility to just getting the microscope to a good-enough zoom level. The task of locating bacilli is non-trivial due to large variances in stain appearance, shape of bacilli clusters, and number of visible bacilli. Moreover, the size of bacilli may differ between slides corresponding to varying zoom levels, which are selected subjectively by the experimenter. We hence choose Deep Convolutional Neural Networks as algorithm for the task, which have shown good results in problems with similar challenges \cite{NIPS2012_4824}.

Deep Learning \cite{deeplearning} refers to Machine Learning algorithms that use different types of multi layered neural networks and have achieved state of the art results on multiple image classification \cite{NIPS2012_4824} and object detection \cite{DBLP:journals/corr/KalinowskiS15}. Like any other Machine Learning solution, we need a dataset with annotations of TB bacteria in smear sputum images to train and test our method. We use dataset made available by \cite{doi:10.1117/1.JMI.4.2.027503} to build a prototype and evaluate our method. We describe the dataset and its usage in our work in detail in the dataset section \ref{Dataset}. Section \ref{Method} describes our Deep Learning based method for TB detection. Section \ref{Results} describes evaluation results of our method's prototype.  The dataset we use is small in comparison to datasets generally used to train deep learning models \cite{imagenet_cvpr09} and thus we can expect to obtain much better results with a larger dataset for TB detection if made available.

\section{Previous Work}
This work has been possible by the release of annotated smear sputum slides made available by \cite{doi:10.1117/1.JMI.4.2.027503} and recent advances in Deep Learning \cite{deeplearning,NIPS2012_4824} on images. To our knowledge, this is the first proof of concept that TB detection can be automated with Deep Learning.
Deep Learning based methods have been able to achieve human accuracy levels in other related fields like cancer metastasis detection \cite{DBLP:journals/corr/LiuGNDKBVTNCHPS17} , Diabetic Retinopathy diagnosis \cite{doi:10.1001/jama.2016.17216} and Skin Cancer diagnosis \cite{NatureThrun}. Patch classification based methods have produced state of the art results in locating anomalies in related domain \cite{MICCAI12Schmidhuber}. We choose to use a patch based classification method for bacillus detection with two-stage cascading. Our neural network architecture is inspired from \cite{DBLP:journals/corr/KalinowskiS15} due to its simplicity.

\label{Dataset}
\section{Dataset}
\begin{figure}[!tpb]
\centerline{
\includegraphics[width=0.5\textwidth]
{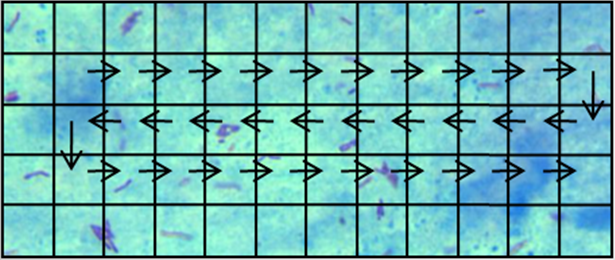}}
\caption{View Fields are sequential overlapping parts of a slide. View-Fields from the same slide are organized as a set in the dataset.}\label{fig:01}
\end{figure}

\begin{figure}[!tpb]
\centerline{
\includegraphics[width=0.5\textwidth]
{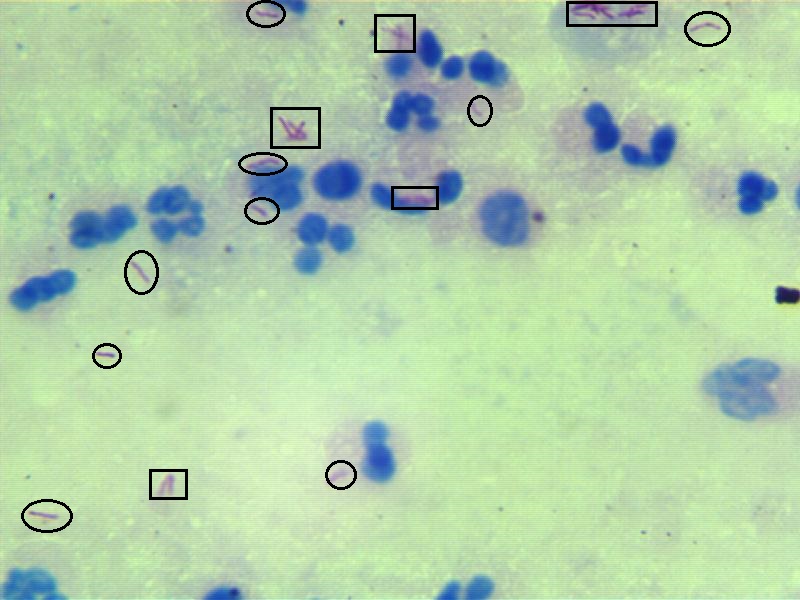}}
\caption{Bacilli marked in dataset 3 are annotated directly on images.}\label{fig:02}
\end{figure}

7 datasets of microscope view-fields made available for different TB diagnostic tasks by \cite{doi:10.1117/1.JMI.4.2.027503} referred to as dataset 1 through 7 respectively. The datasets were obtained from this \href{http://14.139.240.55/znsm/}{url}. A view-field is one part of a slide which is photographed in a way as shown in figure \ref{fig:01}. A set is all view-fields that make a slide. Dataset 3, which has annotations of bacilli bacteria on view-fields, is the dataset meant to train and test a method like ours. However, dataset 3 cannot be directly used for bacillus detection in images as annotations are directly made on images as seen in \ref{fig:02}, which might make the algorithm learn annotation boundary features as discriminating features. Luckily all field-views of dataset 3 are available in dataset 2 as well, without annotations, which helps us to recreate dataset 3 with bounding box coordinates for all bacilli without the annotation boundaries. Another caution we had to take while evaluating our method's prototype on the dataset is being conscious of the fact that consecutive view-fields of a slide have overlapping areas. To avoid a situation where train images and test image might have any common patterns, we use first 60\% consecutive view-fields from all sets as train images and leave the middle 20\% unused and use last 20\% for testing the method. This makes sure that train and test images are exclusive and have no overlap.

\label{Method}
\section{Method}
Our method uses a patchwise detection strategy, that is, for a microscopic field view, 20px X 20px patches from the input image (which is a microscope view-field) are classified one at a time for the presence or absence of bacilli. We use a simple neural network architecture inspired from the neural network architecture in \cite{DBLP:journals/corr/KalinowskiS15}. 

Our architecture is a 5-layered Fully-Convoluted Neural Network Architecture, that is a Convolutional Neural Network \cite{NIPS2012_4824}, without any fully connected units. All non-linearities we use in the network are Rectified Linear Units(ReLUs) except Softmax used in the last layer. In case of cascading, two instances of the same neural network architecture are trained separately.

\begin{figure}[!tpb]
\centerline{
\includegraphics[width=0.5\textwidth]
{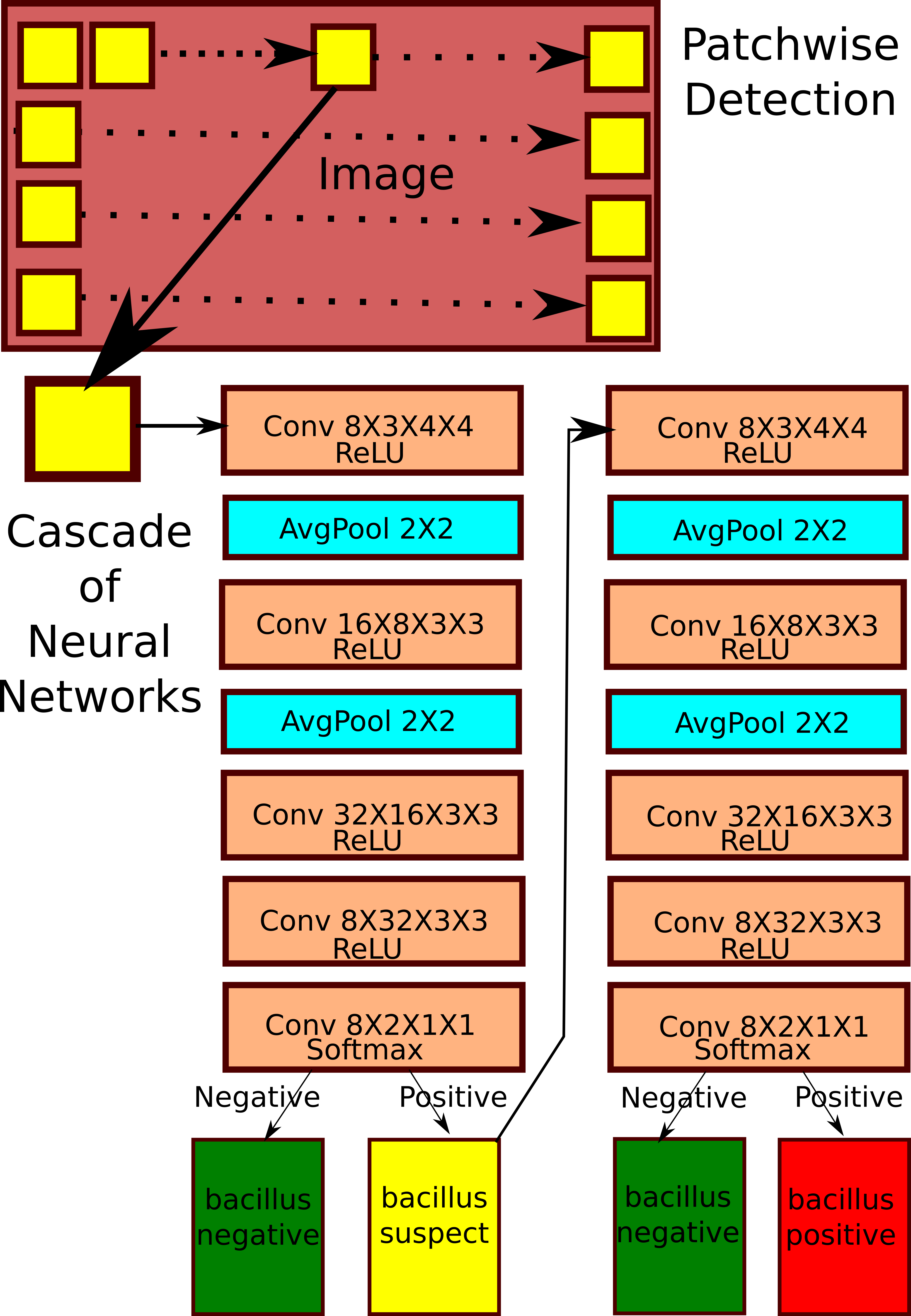}}
\caption{Neural Network Architecture and Cascading.}\label{fig:03}
\end{figure}

The main complexity for building a patch classification based bacillus detector on the dataset is training the classifier on a very skewed class imbalance. The number of bacillus negative patches are many hundred times number of positive patches. The general method in Machine Learning to address class imbalance is undersampling the majority class. Hence, we train a classifier on all positive patches and negative samples randomly sampled such that their number is same as positive patches. The classifier thus trained is biased towards saying most patches as bacillus positive. In statistical terms, this classifier has a large number of type 1 errors (false positives). We thus create a cascade, that is train a second neural network is trained to rectify the errors of initial one trained undersampling the majority class. The second neural network in the cascade is trained to differentiate between true positives and false positives of the first neural network on the train set. \ref{fig:03} shows the neural network architecture and the cascade. 

\label{Results}
\section{Results}
The neural network cascade method we propose achieves a patchwise classification accuracy of 99.8\% on the test set. However, since most of the patches in test images are bacillus negative, we calculate metrics that can evaluate the performance of the algorithm on bacillus positive class. For this we calculate recall(sensitivity), precision(positive predictive value) of the method in locating bacillus bacteria in the test set. Recall (or sensitivity) is the ratio of bacillus positive patches classified correctly to total number of bacillus positive patches in the test images. Precision is the ratio of patches classified as bacillus positive correctly by our method to total number of patches classified as bacillus positive in the test images. The Recall and Precision are calculated to be 83.78\% and 67.55\% respectively. The dataset unfortunately has no slides attributed to patients, hence, it is not possible to ascertain test sensitivity and specificity of diagnosis for our method. That said, since we can detect over 83\% of all bacilli in view-fields, there is a high chance that test sensitivity of the method (ratio of total number of correctly diagnosed patients to total patients) would be high.


\section{Discussion}
With a clear need and mandate to have scalable diagnosis method to check the TB pandemic of India, there should be stress on automation to increase coverage and reducing burden on existing staff who have to assess several million slides an year. Drug-sensitive TB diagnosis in microscopy of sputum smears is essentially a three staged process, involving preparation of slides, zooming in into different parts of a slide and trying to detect bacilli. To make the TB diagnosis process fully automated and fast, all three mentioned steps have to be made automated and fast individually. Our method is an attempt to make the third stage of searching for bacilli in the slides fully automated. We believe this is a tiresome and error-prone task when done by humans and has the potential to be automated totally. Hence the task at a diagnosis site now becomes simpler; putting a slide under the microscope and zooming it to a good-enough level so that bacilli might be visible. As future work, a combined dataset of zoom-levels and bacilli localization might fully automate and fasten diagnosis on the slides.

\section{Conclusion}
We propose a method to automatically detect TB germs in microscope field view with high sensitivity of bacillus detection. We develop a prototype of the method that takes in microscope view-fields with good-enough zoom levels and detects location of bacilli. This system has the potential to automate the TB diagnosis process, hence preventing the spread of this highly contagious disease and one of India's biggest health problems.

{\small
\bibliographystyle{splncs}
\bibliography{egbib}
}

\end{document}